\def\allfiles{}

\documentclass[10pt,twocolumn,letterpaper]{article}

\usepackage[pagenumbers]{cvpr} 

\usepackage{graphicx}
\usepackage{amsmath}
\usepackage[accsupp]{axessibility} 
\usepackage{amssymb}
\usepackage{booktabs}
\usepackage[OT1]{fontenc} 

\usepackage[pagebackref,breaklinks,colorlinks]{hyperref}

\usepackage[capitalize]{cleveref}
\usepackage{multirow}
\crefname{section}{Sec.}{Secs.}
\Crefname{section}{Section}{Sections}
\Crefname{table}{Table}{Tables}
\crefname{table}{Tab.}{Tabs.}

\def\cvprPaperID{742} 
\def\confName{CVPR}
\def\confYear{2023}

\usepackage{color}
\usepackage{CJKutf8}
\definecolor{DeltaColor}{rgb}{0.039,0.73,0.71}
\definecolor{SetaColor}{rgb}{0.867, 0.0235, 0.376}
\definecolor{SigmaColor}{rgb}{0.98,0.45,0.0}
\definecolor{RedColor}{rgb}{0.8,0,0}
\definecolor{AlphaColor}{rgb}{0,0,0.8}
\definecolor{BetaColor}{rgb}{0.8,0,0.8}
\definecolor{GammaColor}{rgb}{0.5,0,0.7}
\definecolor{EpsilonColor}{rgb}{0.353,0.725,0.906}
\definecolor{TauColor}{rgb}{0.423,0.235,0.192}
\definecolor{WtColor}{rgb}{0.235,0.470,0.470}

\begin{document}

\title{



Identity-Preserving Talking Face Generation with\\ Landmark and Appearance Priors
}


\author{Weizhi Zhong$^{1}$\quad\quad Chaowei Fang$^{2}$ \quad\quad Yinqi Cai$^{1}$ \quad\quad Pengxu Wei$^{1}$ \\ Gangming Zhao$^{3}$ \quad\quad  Liang Lin$^{1}$ \quad\quad Guanbin Li$^{1}$\thanks{Corresponding author is Guanbin Li.} \vspace{2mm}\\
$^1$Sun Yat-sen University \quad\quad $^2$Xidian University \quad\quad $^3$The University of Hong Kong\\
{\tt\small \{zhongwzh5, caiyq27\}@mail2.sysu.edu.cn}, {\tt\small chaoweifang@outlook.com} \\{\tt\small \{weipx3, liguanbin, linlng\}@mail.sysu.edu.cn}, {\tt\small gangmingzhao@gmail.com}
	\vspace{-0mm}
}

\maketitle
\begin{abstract}
Generating talking face videos from audio attracts lots of research interest. A few person-specific methods can generate vivid videos but require the target speaker's videos for training or fine-tuning. Existing person-generic methods have difficulty in generating realistic and lip-synced videos while preserving identity information. To tackle this problem, we propose a two-stage framework consisting of audio-to-landmark generation and landmark-to-video rendering procedures. First, we devise a novel Transformer-based landmark generator to infer lip and jaw landmarks from the audio. Prior landmark characteristics of the speaker's face are employed to make the generated landmarks coincide with the facial outline of the speaker. Then, a video rendering model is built to translate the generated landmarks into face images. During this stage, prior appearance information is extracted from the lower-half occluded target face and static reference images, which helps generate realistic and identity-preserving visual content. For effectively exploring the prior information of static reference images, we align static reference images with the target face's pose and expression based on motion fields. Moreover, auditory features are reused to guarantee that the generated face images are well synchronized with the audio. Extensive experiments demonstrate that our method can produce more realistic, lip-synced, and identity-preserving videos than existing person-generic talking face generation methods. Project page: \url{https://github.com/Weizhi-Zhong/IP_LAP}
\end{abstract}



\section{Introduction}
\label{sec:intro}
\begin{figure}[t]
  \centering
  \includegraphics[width=\linewidth]{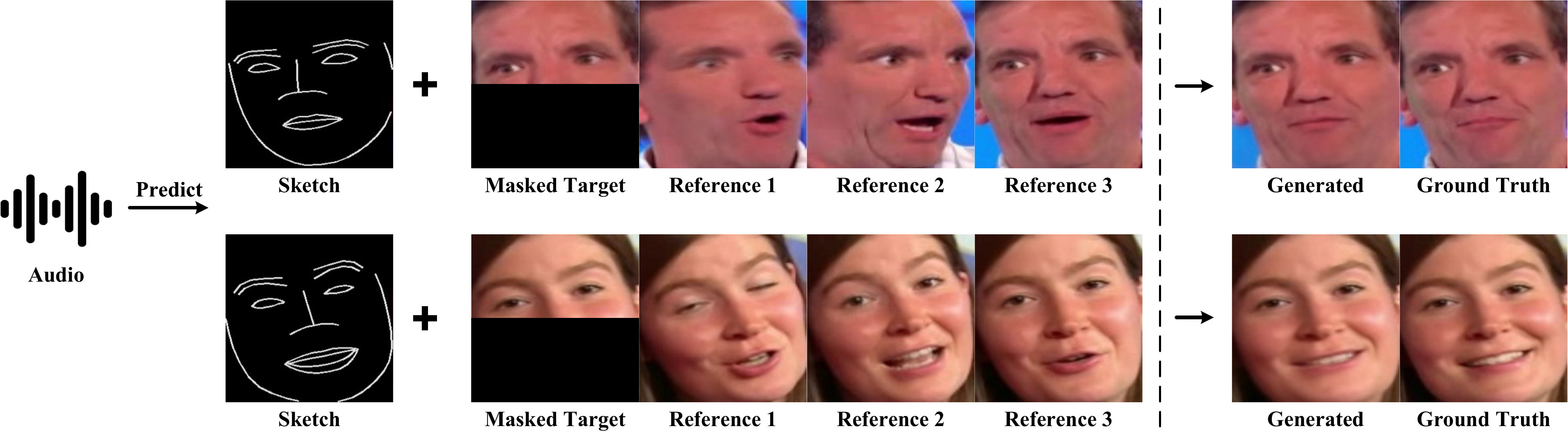}
  \vspace{-17pt}
  \caption{This paper is targeted at generating a talking face video for a speaker which is coherent with input audio.
  We implement this task by completing the lower-half face of the speaker's original video. The outline of the mouth and jaw is inferred from the input audio and then used to guide the video completion process. Moreover, multiple static reference images are used to supply prior appearance information.
   }
   \vspace{-15pt}
  \label{fig:intro}
\end{figure}

Audio-driven talking face video generation is valuable in a wide range of applications, such as visual dubbing\cite{kr2019towards,prajwal2020lip,xie2021towards}, digital assistants\cite{thies2020neural}, and animation movies\cite{zhou2020makelttalk}. 
Based on the training paradigm and data requirement, the talking face generation methods can generally be categorized as person-specific or person-generic types.
Person-specific methods\cite{thies2020neural,liu2022semantic,lu2021live,shen2022learning,zhang2021facial,guo2021ad} can generate photo-realistic talking face videos but need to be re-trained or fine-tuned with the target speaker's videos, which might be inaccessible in some real-world scenarios.
Hence, learning to generate person-generic talking face videos is a more significant and challenging problem in this field. This topic also attracts lots of research attention~\cite{prajwal2020lip,kr2019towards,xie2021towards,park2022synctalkface,zhou2021pose,huang2022audio}. 
In this paper, we focus on tackling the person-generic talking face video generation by completing the lower-half face of the speaker's original video under the guidance of audio data and multiple reference images, as shown in Figure~\ref{fig:intro}. 

The main challenges of the person-generic talking face video generation include two folds: 1) How can the model generate videos having facial motions, especially mouth and jaw motions, which are coherent with the input audio? 2) How can the model produce visually realistic frames while preserving the identity information?
To address the first problem, many methods~\cite{ji2021audio,xie2021towards,biswas2021realistic,chen2020talking,chen2019hierarchical,zhou2020makelttalk} leverage facial landmarks as intermediate representation when generating person-generic talking face videos. 
However, translation from audio to facial landmarks is an ambiguous task, considering the same pronunciation may correspond to multiple facial shapes. A few landmark-based talking face generation methods~\cite{zhang2020apb2face,biswas2021realistic} tend to produce results having the averaged lip shape of training samples, which may have remarkable differences with the lip shape of the speaker.
A line of methods~\cite{chen2019hierarchical,xie2021towards} incorporate the prior information from a reference image's landmarks to generate landmarks consistent with the speaker's shape. 
However, they directly fuse the features of audio and landmarks with simple concatenation or addition operations without modeling  the underlying correlation between them. 
For example, the relations between the reference landmarks and the audio clips from different time intervals are different. Those methods are not advantageous at capturing such kinds of differences. 
Moreover, the temporal dependencies are also valuable for predicting facial landmarks. 
Existing methods, such as~\cite{yi2022animating,ji2021audio,zhou2020makelttalk,chen2019hierarchical}, depend on long short-term memory (LSTM) models to explore the temporal dependencies when transforming audio clips to landmark sequences. However, those models are limited in capturing 
long-range temporal relationships.
%


Since the input audio and intermediate landmarks do not contain visual content information intrinsically, it is very challenging to hallucinate realistic facial videos from audio and intermediate landmarks while preserving the identity information.
A few existing methods, such as~\cite{xie2021towards,zhou2020makelttalk}, adopt a single static reference image to supply visual appearance and identity information. 
However, one static reference image is insufficient to cover all facial details, e.g., the teeth and the side content of cheeks. This makes these algorithms struggle to synthesize unseen details, which is unreliable and easily leads to generation artifacts.
\cite{chung2017you,biswas2021realistic} use multiple reference images to provide more abundant details. 
However, they simply concatenate the reference images without spatial alignment, which is limited in extracting meaningful features from reference images. 

To cope with the above problems, we devise a novel two-stage framework composed of an audio-to-landmark generator and a landmark-to-video rendering network. 
The goal of our framework is to complete the lower-half face of the video with content coherent to the phonetic motions of the audio. 
Specifically, we use pose prior landmarks of the upper-half face and reference landmarks extracted from static face images as extra inputs of the audio-to-landmark generator.
The access to the two kinds of landmarks helps to prevent the generator from producing results that deviate from the face outline of the speaker.
Then, we build up the network architecture of the generator based on the multi-head self-attention modules~\cite{vaswani2017attention}. Our design is more advantageous at capturing relationships between phonetic units and landmarks compared to simple concatenation or addition operations~\cite{chen2019hierarchical,xie2021towards}. It is also more helpful for modeling temporal dependencies than LSTM used in previous methods~\cite{yi2022animating,ji2021audio,zhou2020makelttalk,chen2019hierarchical}. 
Additionally, multiple static face images are referred to extract prior appearance information for generating realistic and identity-preserving face frames. 
Inspired by~\cite{doukas2022free}, we set up the landmark-to-video rendering network with a motion field based alignment module and a face image translation module. 
The alignment module is targeted at registering static reference images with the face pose and expression delivered by the results of the landmark generator. 
This target is achieved by inferring a motion field for each static reference image and then warping the image and its features. 
The alignment module can decrease the difficulty in translating meaningful features of static reference images to the target image.
The face image translation module produces the final face images by combining multi-source features from the inferred landmarks, the occluded original images, the registered reference images, and the audio. 
The inferred landmarks provide vital clues for constraining the facial pose and expression. Those images are paramount for inferring the facial appearance. Besides, the audio features are reused to guarantee that the generated lip shapes are well synchronized with the audio. 
Extensive experiments demonstrate that our method produces more realistic and lip-synced talking face videos and preserves the identity information better than existing methods.
Our main contributions are summarized as follows:
\begin{itemize}
    \item We propose a two-stage framework composed of an audio-to-landmark generator and a landmark-to-video rendering model to address the person-generic talking face generation task under the guidance of prior landmark and appearance information.
    
    \item We devise an audio-to-landmark generator that  can effectively fuse prior landmark information with the audio features. We also make an early effort to construct the generator with multi-head self-attention modules.
    \item We design a landmark-to-video rendering model which can make full use of multiple source signals, including prior visual appearance information, landmarks, and auditory features.
    \item Extensive experiments are conducted on LRS2~\cite{Afouras18c} and LRS3~\cite{Afouras18d} dataset, demonstrating the superiority of our method over existing methods in terms of realism, identity preservation, and lip synchronization. 
\end{itemize}
\ifx\allfiles\undefined

\input{macros}

\begin{document}
\fi

\section{Related Work}


\subsection{Audio-Driven Talking Face Generation}
Existing audio-driven talking face generation methods mainly include two types, person-specific and person-generic methods.
With the help of 3D Morphable Model (3DMM)~\cite{blanz1999morphable} and  neural radiance fields (NeRF)~\cite{mildenhall2021nerf}, some person-specific methods \cite{thies2020neural,zhang2021facial,zhang2022meta,yi2022predicting,liu2022semantic,shen2022learning,guo2021ad} can synthesize high-fidelity talking face videos. 
For example, NVP \cite{thies2020neural} and FACIAL \cite{zhang2021facial} first predict 3DMM expression parameters from audio and then use a neural rendering network to produce videos.
A few works~\cite{guo2021ad,liu2022semantic,shen2022learning} attempt to implement the audio-driven talking face video generation by controlling the dynamic neural radiance field with the audio and then rendering face images.
Nevertheless, all these methods require videos of the target speaker for re-training or fine-tuning, which may be inaccessible in some real-world scenarios. 
It is more significant to devise person-generic methods which can synthesize talking face videos for unseen speakers. 
There exists a series of literature focusing on person-generic talking face generation, such as~\cite{xie2021towards,chen2019hierarchical,zhou2020makelttalk,10.1145/3528233.3530745,prajwal2020lip,park2022synctalkface,zhou2021pose,wang2022one}.  
Typically, Wav2Lip~\cite{prajwal2020lip} uses an encoder-decoder based generator to synthesize talking face videos under the guidance of a lip sync discriminator. 
Based on Wav2Lip, SyncTalkFace~\cite{park2022synctalkface} proposes an audio-lip memory to provide extra visual information of the mouth region. 
PC-AVS~\cite{zhou2021pose} modularizes talking faces into feature spaces of speech content, head pose, and identity, respectively. These features are assembled to produce talking face videos. 
However, these methods are insufficient in generating highly realistic and lip-synced videos while preserving identity information.


\subsection{Landmark-based Talking Face Generation}
Many audio-driven talking face generation methods \cite{ji2021audio,xie2021towards,zhang2020apb2face,chen2020talking,chen2019hierarchical,zhou2020makelttalk,suwajanakorn2017synthesizing,yi2022animating,tang2021talking,biswas2021realistic} use facial landmarks as intermediate representation.
For example, Suwajanakorn \etal \cite{suwajanakorn2017synthesizing} use a recurrent neural network (RNN) to learn the mapping from audio input to mouth landmarks and then synthesize high-quality mouth texture.
\cite{ji2021audio,lu2021live,chen2019hierarchical,tang2021talking} rely on Long Short Term Memory (LSTM) models to learn the mapping from audio to landmark movements.
MakeItTalk\cite{zhou2020makelttalk} combines LSTM and self-attention mechanisms to predict the landmark displacement from audio. 
APB2Face\cite{zhang2020apb2face} uses linear layers to predict landmark geometry from the concatenated pose, blink, and audio features. 
Xie \etal \cite{xie2021towards} devise a two-stage framework where three encoders are used to extract the audio, pose, and reference embeddings in the first stage, and three embeddings are fused with trainable weights and fed into a decoder to predict the landmarks of the lip and jaw in the second stage.
Our method is also composed of two stages, including audio-to-landmark generation and landmark-to-video rendering. 
It is distinct from existing methods in the following aspects. First, built upon Transformer modules, the audio-to-landmark generator is more advantageous at exploring prior landmark information to predict facial landmarks accurately.
Secondly, the landmark-to-video rendering model can effectively combine multi-source features from prior appearance information, landmarks, and audio, achieving better performance in terms of realism, lip synchronization, and identity preservation than existing methods.

\ifx\allfiles\undefined
{\small
\bibliographystyle{ieee_fullname}
\bibliography{egbib.bib}
}
\end{document}
\fi
\ifx\allfiles\undefined

\input{macros}

\begin{document}
\fi

\begin{figure*}[t]
  \centering 
  \includegraphics[width=\linewidth]{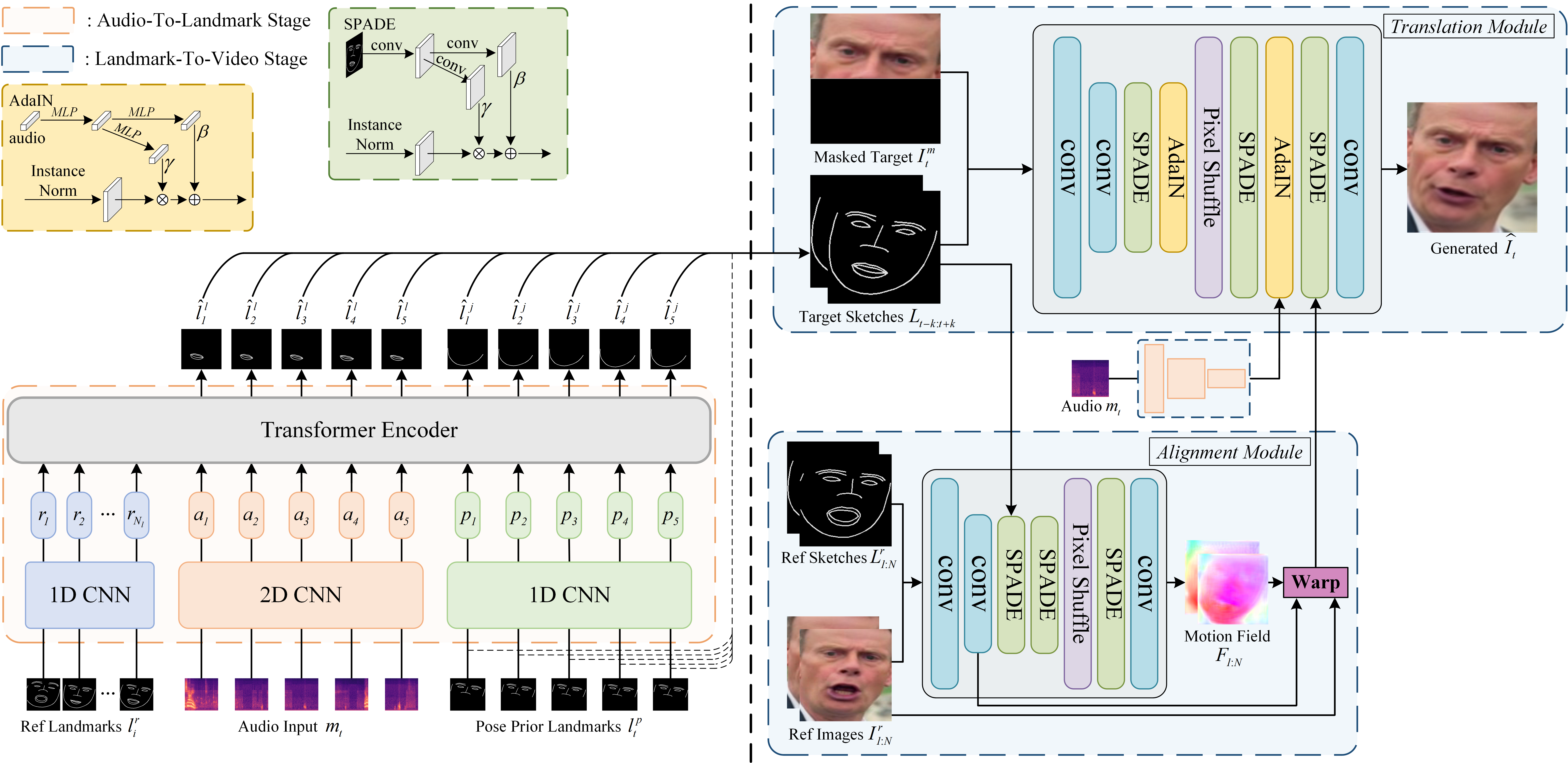}
  \caption{Overview of our framework.~It can be divided into two stages: \textbf{(1) Audio-To-Landmark Generation} (left orange part). The transformer-base landmark generator takes the audio, reference landmarks, and pose prior landmarks as input to predict the landmarks of lip and jaw, which are then combined with pose prior landmarks to construct the target sketches. Positional encodings and modality encodings are omitted for simplicity. \textbf{(2) Landmark-To-Video Rendering} (right blue part). According to target sketches, the alignment module takes multiple reference images and their sketches as input to obtain the motion fields, which warp the reference images and their features to target head pose and expression. With the assistance of audio features and warped images and features, the translation module translates the target sketches concatenated with the lower-half masked target face to the resulted face image.
}
  \label{fig:framework}
\end{figure*}

\section{Proposed Method}  
Given an audio sequence and an initial input video, we aim to generate a lip-synced talking face video by completing the lower-half occluded face of the input video in a frame-by-frame manner.
An overview of our method is depicted in Figure \ref{fig:framework}. 
Our framework is composed of two stages.
The first stage takes the audio signal and prior landmarks of the speaker's face as input to predict the landmarks of the lip and jaw.
The second stage consists of an alignment module and a translation module. 
Based on the motion fields, the alignment module registers reference images and their features with the target face pose and expression. 
The translation module synthesizes the full face image from landmarks under the guidance of auditory features and prior appearance information from the occluded target face and registered reference images. 

\subsection{Audio-To-Landmark Generation}
In this stage, the network aims to generate landmarks of lip  $\{\hat{l}_t^l\in \mathbb{R}^{2\times n_l}\}_{t=1}^T$ and jaw $\{\hat{l}_t^j\in \mathbb{R}^{2\times n_j}\}_{t=1}^T$ for $T=5$ adjacent frames at a time, given the reference landmarks $\{l_i^r\in \mathbb{R}^{2\times n_r}\}_{i=1}^{N_l}$, pose prior landmarks $\{l_t^p\in \mathbb{R}^{2\times n_p}\}_{t=1}^T$, and audio input $\{m_t\in \mathbb{R}^{h\times w}\}_{t=1}^T$. 
$N_l$ is the number of reference landmarks (measured in frames). 
$n_l$, $n_j$, $n_r$, and $n_p$ is the number of landmarks representing lip, jaw, whole face, and pose, respectively.
Following \cite{prajwal2020lip}, for each video frame, we extract out the corresponding audio interval and process it to Mel-spectrogram with size of $16\times 80$, \ie, $h=16$ and $w=80$. 
The reference landmarks provide prior personalized facial outline information for landmark prediction. 
\subsubsection{Transformer-based Landmark Generator}
We extract the audio embedding $a_t$ from the audio Mel-spectrograms with an encoder module $E_a$ similar to \cite{prajwal2020lip}. 
We also utilize 1D convolution layers to construct a pose encoder $E_p$ and a reference encoder $E_r$ which extract pose embedding $p_t$ and reference embedding $r_i$ from pose prior landmarks $l_t^p$ and reference landmarks $l_i^r$, respectively.
This process is formulated as follows: 
\begin{align}
    & a_t=E_a\left (m_t\right )\quad  t=1,2,...,T \\
    & p_t=E_p\left(l_t^p\right)\quad t=1,2,...,T  \\
    & r_i=E_r\left (l_i^r\right ) \quad i=1,2,...,N_l
\end{align}
where $a_t,p_t,r_i\in \mathbb{R}^{d}$ ($d$ is the dimension of audio, pose, or reference embedding). 

For differentiating embeddings calculated from there types of source signals, we introduce three learnable encoding vectors $e_{a}^{type}$, $e_{p}^{type}$, and $e_{r}^{type}\in\mathbb{R}^{d}$, indicating that the embedding is calculated from audio Mel-spectrogram, prior pose landmarks, and reference landmarks, respectively.
The temporal positional encoding of the $t$-th frame is denoted as $e_t^{pos}\in \mathbb{R}^{d}$ which is calculated following the sinusoidal positional encoding.~%
These encoding variables are added to the audio, pose, and reference embeddings as below,
\begin{align}
    & \overline{a_t}=a_t+e_t^{pos}+e_a^{type} &t=1,2,...,T \\
    & \overline{p_t}=p_t+e_t^{pos}+e_p^{type} &t=1,2,...,T  \\
    & \overline{r_i}=r_i+e_r^{type} &i=1,2,...,N_l 
\end{align}

Afterwards, we employ the Transformer modules~\cite{vaswani2017attention}  to capture both intra-type and inter-type relation among three types of embeddings.~The initial tokens $z^0$ are formed by concatenating $\{\overline{r_i}\}_{i=1}^{N_l}$, $\{\overline{a_t}\}_{t=1}^T$, and $\{\overline{p_t}\}_{t=1}^T$. 
Practically, $L$ Transformer encoder modules are adopted in our model, and each module is constituted by a stack of multi-head self-attention (MSA), layer normalization (LN), and MLP layers.
The calculation process is summarized as below,
\begin{equation}
\begin{aligned}
\overline{z}^{\ell} &=\operatorname{MSA}(\operatorname{LN}(z^{\ell-1}))+z^{\ell-1}, & \ell &=1 \ldots L \\
z^{\ell} &=\operatorname{MLP}(\operatorname{LN}(\overline{z}^{\ell}))+\overline{z}^{\ell}, & \ell &=1 \ldots L
\end{aligned}
\end{equation}
where $z^{\ell}\in \mathbb{R}^{(N_l+2T)\times d}$ represents the output of the $\ell$-th Transformer module.
Let $z_i^L$ be the representation of the $i$-th token in $z^L$.
The second last $T$ tokens are used to predict  lip landmarks, and the last $T$ tokens are used to predict jaw landmarks:
\begin{align}
    &\hat{l}_t^l=\operatorname{MLP}(z_{t+N_l}^{\textit{L}}) &t=1,2,...,T\\
    &\hat{l}_t^j=\operatorname{MLP}(z_{t+N_l+T}^{\textit{L}}) &t=1,2,...,T
\end{align}
where $\hat{l}_t^l\in \mathbb{R}^{2\times n_l}$ and $\hat{l}_t^j\in \mathbb{R}^{2\times n_j}$ denote the predicted landmarks of lip and jaw at the $t$-th frame, respectively.

\subsubsection{Loss Function for Landmark Generation}
During training, we first apply the L1 reconstruction loss $L_1$ to constrain the predicted landmarks:
\begin{equation}
    L_{1} = \frac{1}{T} \sum_{t=1}^{T}\left ( \|\hat{l}_t^l-l_t^l\|_{1}+\|\hat{l}_t^j-l_t^j\|_{1}\right )
\end{equation}
where $l_t^l$ and $l_t^j$ denote the ground-truth landmarks of lip and jaw, respectively.

For sake of improving the temporal smoothness, we also adopt the following continuity regularization to constrain the predicted landmarks,
\begin{equation}
    \begin{split}
    L_{c} = \frac{1}{T-1}\sum_{t=1}^{T-1}&\left(\|(\hat{l}_{t+1}^l-\hat{l}_t^l)-(l_{t+1}^l-l_t^l)\|_{2} \right.\\
 &\left.+\|(\hat{l}_{t+1}^j-\hat{l}_t^j)-(l_{t+1}^j-l_t^j)\|_{2}\right) 
    \end{split}
\end{equation}
The overall training loss for audio-to-landmark stage is defined as follows:
\begin{equation}
    L = L_1+ \lambda_c L_c
\end{equation}
where $\lambda_c$ is a constant and is set to 1.
\subsection{Landmark-To-Video Rendering}
Inspired from~\cite{doukas2022free}, we design a rendering stage consisting of an alignment module $G_a$ and a translation module $G_r$.
At first, the predicted landmarks are assembled with the pose prior landmarks, forming a complete set of facial landmarks. Then we draw these landmarks on the image plane to get the target face sketches. 
We denote the target face sketch at the $t$-th frame as $L_t$.
To achieve temporal continuity across frames, we select $2k+1$ target sketches around the $t$-th frame, $\{L_i \in \mathbb{R}^{3\times H \times W}\}_{i=t-k}^{t+k}$ as inputs for predicting the $t$-th face image. 
Besides, to provide more appearance prior for the realistic rendering, multiple reference images $\{I_i^r\in \mathbb{R}^{3\times H\times W}\}_{i=1}^N$ and their 
extracted sketches $\{L_i^r\in \mathbb{R}^{3\times H\times W}\}_{i=1}^N$ are input to the alignment module for computing the motion fields $\{F_i\in \mathbb{R}^{2\times H\times W}\}_{i=1}^N$. The reference images and their features are then warped to target head pose and expression by the motion fields.  
The translation module then translates the target sketches into the final face image $\hat{I}_t\in \mathbb{R}^{3\times H\times W}$, under the guidance of auditory features, warped reference images and features, and the lower-half occluded target face. 
\subsubsection{Reference Images Warping}
For each reference image $I_i^r$ and its sketch $L_i^r$, the alignment module encodes the channel-wise concatenation of $L_i^r$ and $I_i^r$ with convolution layers to visual features in two spatial resolutions: $h_i^1\in \mathbb{R}^{c_1\times h_1\times w_1}, h_i^2 \in \mathbb{R}^{c_2\times h_2\times w_2} $. 
To guide the prediction of motion field $F_i$, the channel-wise concatenation of $2k+1$ target sketches $L_{t-k:t+k}$ are injected into the alignment module by SPADE\cite{park2019semantic} layers, which modulate the visual features according to target sketches.
Here, PixelShuffle layers\cite{shi2016real} are utilized for up-sampling. Generally, the function of the alignment module can be formulated as:
\begin{equation}
    F_i=G_a(L_i^r,I_i^r,L_{t-k:t+k}) \quad i=1,2,...,N 
\end{equation}
Besides, for the aggregation of multiple reference images and corresponding visual features warped by the motion fields, one more output layer is added to the alignment module to predict a 2D weight $w_i\in \mathbb{R}^{H\times W}$ for $I_i^r$. The aggregated warped image can be calculated as:
\begin{align}
    \overline{I^r}=\frac{\sum_{i=1}^{N}  w_i F_i(I_{i}^{r})}{\sum_{i=1}^{N}  w_{i}} 
\end{align}
where  $F_i(I_{i}^{r})\in \mathbb{R}^{3\times H\times W}$ is the image warped by the motion field $F_i$. And the aggregated warped features in two spatial resolutions are $\overline{h}^{1}$ and $\overline{h}^{2}$, respectively, which are calculated as follows:
\begin{align}
    \overline{h}^{s}=\frac{\sum_{i=1}^{N} w_{i} F_i(h_i^s)}{\sum_{i=1}^{N}  w_i} \quad s=1,2
\end{align}
where $F_i(h_i^s)\in \mathbb{R}^{c_s\times h_s\times w_s}$ is the visual feature warped by the motion field $F_i$. 
Note that $F_i$ and $w_i$ are downsampled to match the size of $h_i^s$. 

\subsubsection{Sketch-To-Face Translation}

In the translation module $G_r$ , we aim to translate the target sketches concatenated with the masked target face $I_t^m$ to the final face image $\hat{I}_t$. 
This is performed via the assistance of aggregated warped image $\overline{I^r}$ and feature  $\overline{h}^1, \overline{h}^2$. 
Besides, to enhance the synthesized mouth detail and lip synchronization, the auditory feature $a_t\in \mathbb{R}^{d}$ extracted by an audio encoder similar to \cite{prajwal2020lip} is injected into the translation module through AdaIN layers\cite{huang2017arbitrary}.
The overall process can be formulated as:
\begin{equation}
   \hat{I}_t=G_r(I_t^m,L_{t-k:t+k},\overline{I^r},\overline{h}^1,\overline{h}^2,a_t) 
\end{equation}
Specifically, $I_t^m$ and $L_{t-k:t+k}$ are concatenated channel-wise and fed into the convolution layers to obtain encoded features.  $\overline{I^r}$, $\overline{h}^1$, and $\overline{h}^2$ are fused into the translation module through SPADE\cite{park2019semantic} layers to modulate the encoded features followed by AdaIN operation\cite{huang2017arbitrary}. Meanwhile, up-sampling is implemented through PixelShuffle layers\cite{shi2016real,yue2021robust} for better performance. 

\begin{figure}[hb]
  \centering
  \includegraphics[width=0.9\linewidth]{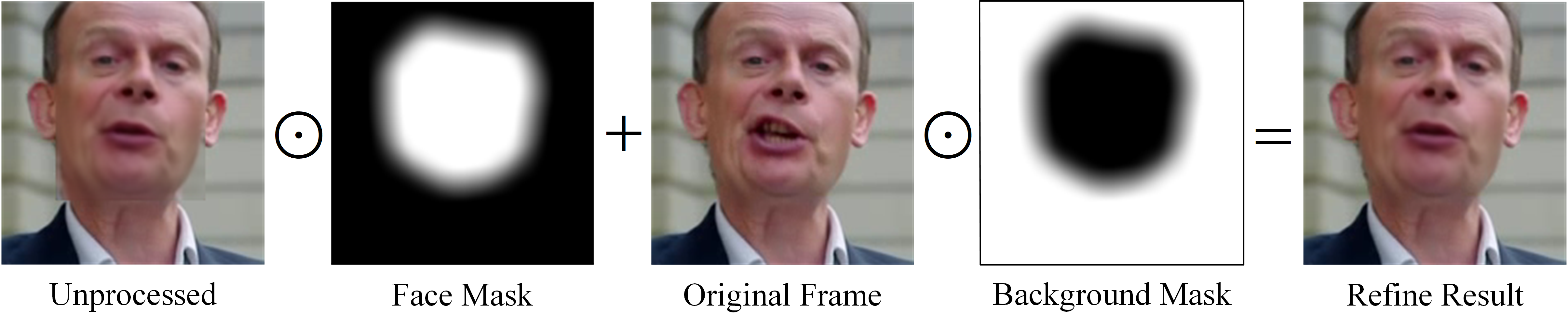}
  \caption{Post-processing for generated face pasted onto the original frame through a Gaussian-smoothed face mask.  }
  \label{fig:post}
  \vspace{-5pt}
\end{figure}

At last, the generated full face is pasted onto the original frame during inference. However, since the generated face may include small portions of background with artifacts, we composite the generated face with the background of the original frame through a Gaussian-smoothed face mask, as depicted in Figure \ref{fig:post}.
\subsubsection{Loss Function for Rendering}
Following~\cite{ren2021pirenderer}, the alignment module, translation module, and audio encoder are trained jointly on the reconstruction task. The warping loss $L_w$, based on the perceptual loss of~\cite{johnson2016perceptual}, between aggregated warped image $\overline{I^r}$ and the ground truth image $I_t$, is employed to constrain the alignment module for computing an accurate motion field. 
\begin{equation}
    L_{w}=\sum_{i}\left\|\phi_{i}(\overline{I^r})-\phi_{i}(I_t)\right\|_{1}
\end{equation}
where $\phi_{i}$ is the activation output of the $i$-th layer in the VGG-19 network. 
The reconstruction loss $L_r$, which has a similar structure as that of $L_w$, and the style loss $L_s$, which computes the statistic error between activation output in the VGG-19 network, are applied to reduce errors between generated face $\hat{I}_t$ and ground truth $I_t$.
\begin{align}
    L_{r} &=\sum_{i}\left\|\phi_{i}(\hat{I}_t)-\phi_{i}(I_t)\right\|_{1}  \\
    L_{s} &=\sum_{i}\left\|G_{i}^{\phi}(\hat{I}_t)-G_{i}^{\phi}(I_t)\right\|_{1}
\end{align}
where $G_{i}^{\phi}$ is the Gram matrix derived from the aforementioned activation output $\phi_{i}$.
To further enhance the  photo-realism of the generated image, we also utilize patch GAN loss $L_g$ and feature matching loss $L_f$ from pix2pixHD\cite{wang2018high}. To summarize, the training loss for the landmark-to-video rendering stage can be formulated as follows:
\begin{equation}
    L = \lambda_w L_w+ \lambda_r L_r+ \lambda_s L_s + \lambda_g L_g+ \lambda_f L_f
\end{equation}
In experiments, we set $\lambda_w=2.5$, $\lambda_r=4$, $\lambda_s=1000$, $\lambda_g=0.25$, and $\lambda_f=2.5$ based on cross-validation. 

\ifx\allfiles\undefined
{\small
\bibliographystyle{ieee_fullname}
\bibliography{egbib.bib}
}
\end{document}
\fi





\begin{table*}[t]
\fontsize{8}{10} \selectfont
\centering
 \setlength{\tabcolsep}{2mm}{
\begin{tabular}{c|c|cccccc|c}
\hline
\multirow{2}{*}{\textbf{Method}} & \multirow{2}{*}{\textbf{Dataset}} & \multicolumn{6}{c|}{\textbf{Reconstruction}}                                                             & \textbf{Dubbing}    \\ \cline{3-9} 
                                 &                                   & \textbf{PSNR$\uparrow$} & \textbf{SSIM$\uparrow$}  & \textbf{LPIPS$\downarrow$} & \textbf{FID$\downarrow$}  & \textbf{LipLMD$\downarrow$} & \textbf{CSIM$\uparrow$}  & \textbf{SyncScore$\uparrow$} \\ \hline

ATVGnet \cite{chen2019hierarchical}  &    \multirow{6}{*}{LRS2}        & 11.55          & 0.3944          & 0.5575          & 223.26         & 0.27401          & 0.1020          & 3.51                \\

Wav2lip \cite{prajwal2020lip}                         &                                   & 27.92          & 0.8962          & 0.0741          & 43.46          & 0.02003          & 0.5925          & 3.86                \\

MakeItTalk \cite{zhou2020makelttalk}                      &                                   & 17.25          & 0.5562          & 0.2237          & 76.57          & 0.05024          & 0.5799          & 2.68                \\

PC-AVS \cite{zhou2021pose}                           &                                   & 15.75          & 0.4867          & 0.2802          & 110.60         & 0.07569          & 0.3927          & \textbf{5.20}       \\

EAMM \cite{10.1145/3528233.3530745}                            &                                   & 15.17          & 0.4623          & 0.3398          & 91.95          & 0.15191          & 0.2318          & 3.01                \\

\textbf{Ours}                    &                                   & \textbf{32.91} & \textbf{0.9399} & \textbf{0.0303} & \textbf{27.87} & \textbf{0.01293} & \textbf{0.6523} & 4.49                \\ \hline

ATVGnet \cite{chen2019hierarchical}      & \multirow{6}{*}{LRS3}          & 10.90          & 0.3791          & 0.5667          & 190.49         & 0.30564          & 0.1176          & 4.43                \\

Wav2lip \cite{prajwal2020lip}                          &                                   & 28.45          & 0.8852          & 0.0683          & 49.60          & 0.02001          & 0.5909          & 4.26                \\

MakeItTalk \cite{zhou2020makelttalk}                       &                                   & 17.78          & 0.5607          & 0.2788          & 97.99          & 0.08432          & 0.5465          & 3.04                \\

PC-AVS \cite{zhou2021pose}                           &                                   & 15.60          & 0.4732          & 0.3321          & 115.25         & 0.10611          & 0.3537          & \textbf{6.15}       \\

EAMM \cite{10.1145/3528233.3530745}                             &                                   & 15.37          & 0.4679          & 0.3812          & 108.83         & 0.17818          & 0.2689          & 3.30                \\

\textbf{Ours}                    &                                   & \textbf{32.97} & \textbf{0.9222} & \textbf{0.0310} & \textbf{29.96} & \textbf{0.01353} & \textbf{0.6385} & 5.63                \\ \hline

\end{tabular}%
\vspace{-5pt}
}
\caption{Quantitative comparison with state-of-the-art person-generic talking face generation methods on the reconstruction and dubbing setting. $\uparrow$ indicates higher is better while $\downarrow$ indicates lower is better.}
\label{tab:table1}
\end{table*}

\section{Experiments}    
\subsection{Experimental Settings}
\textbf{Implementation Details.}  
When generating landmarks from audio, we utilize the mediapipe tool~\cite{lugaresi2019mediapipe} to detect facial landmarks from each video frame, where $n_l=41$, $n_j=16$, $n_r=131$, and $n_p=74$. 
Following \cite{prajwal2020lip}, we calculate Mel-spectrograms from 16kHz audios using a window size of 800 and hop size of 200. $N_l$ is set to 15, and $d$ is 512. 
To render videos from landmarks,  $128\times128$ face images (i.e., $H=W=128$) are generated at 25 fps, and $k$ is set to 2. 
The number of reference images $N$ is set to 3 during training and $1/5$ of the video length during inference.
More details about network architectures and hyper-parameters are included in the supplementary document.
\vspace{3mm}

\textbf{Dataset.} Two audio-visual speech recognition datasets, LRS2 \cite{Afouras18c} and LRS3 \cite{Afouras18d}, are used in our experiments. 
\begin{itemize}
    \item[1)] \textbf{LRS2} \cite{Afouras18c}. The dataset consists of 48,164 video clips from outdoor shows on BBC television. Each video is accompanied by an audio corresponding to a sentence with up to 100 characters. The training, validation, and test sets are split based on broadcast date, including 45,839, 1,082, and 1,243 videos, respectively. We sample 45 videos from the test set for evaluating algorithms quantitatively. 
    \item[2)] \textbf{LRS3} \cite{Afouras18d}. This dataset consists of 151,819 videos from indoor shows of TED or TEDx.  There exists no overlap between LRS2 and LRS3. The videos of the two datasets are distinct in shooting scenes, lighting conditions, actions, etc. Thus, we sample 45 videos from LRS3 to test the generalization ability of  talking face generation methods.
\end{itemize}
\vspace{3mm}

\begin{figure*}[ht]
\vspace{-5pt}
  \centering
  \includegraphics[width=0.75\linewidth]{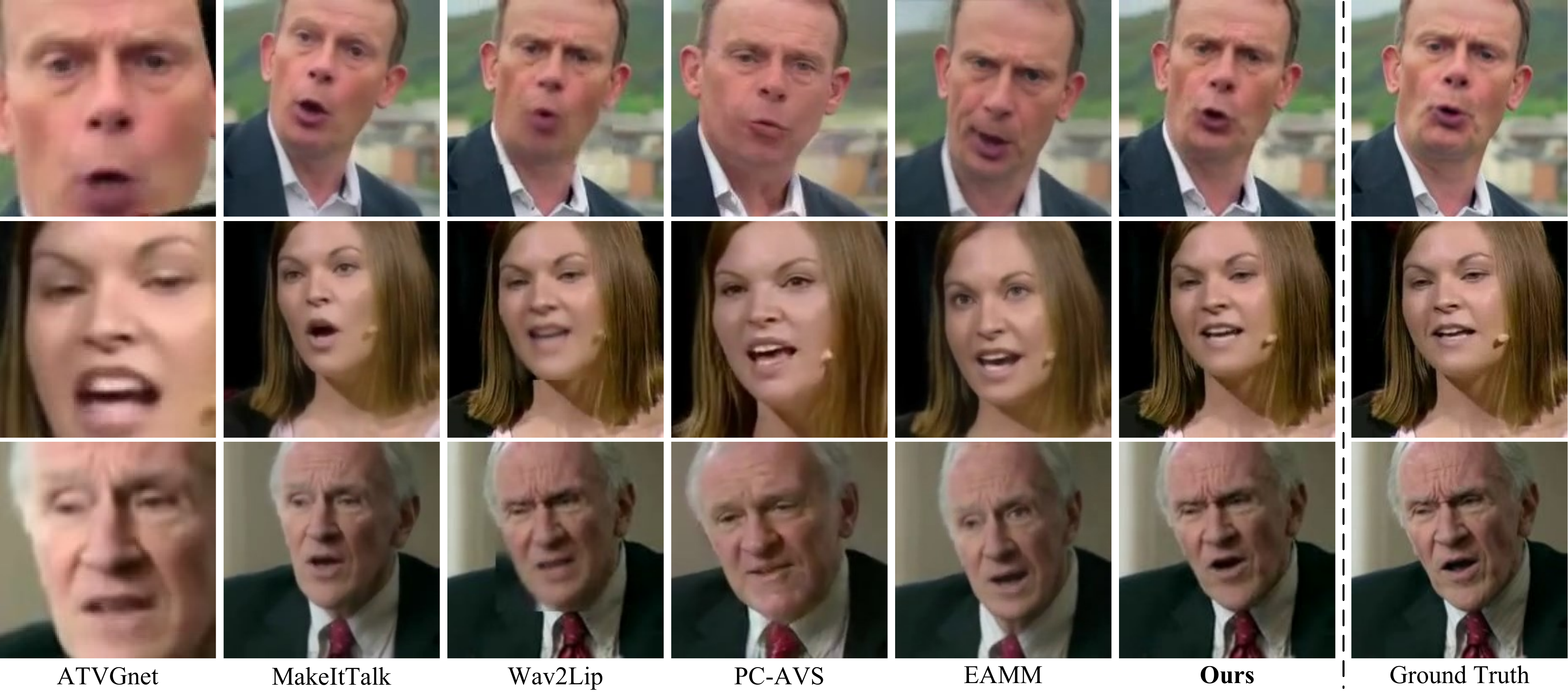}
  \vspace{-5pt}
  \caption{Qualitative comparisons with state-of-the-art person-generic methods on LRS2 \cite{Afouras18c} and LRS3 \cite{Afouras18d} datasets. Our method produces realistic results with better identity preservation effect. More results are presented in the supplementary video.
}
\vspace{-13pt}
  \label{fig:Qualitative}
\end{figure*}

\textbf{Comparison Methods.}
We compare our method against state-of-the-art methods~\cite{chen2019hierarchical,zhou2020makelttalk,prajwal2020lip,zhou2021pose,10.1145/3528233.3530745} on the person-generic audio-driven talking face generation. 
\textbf{EAMM}~\cite{10.1145/3528233.3530745} generates talking face videos with emotion control based on unsupervised motion representations.
\textbf{PC-AVS}~\cite{zhou2021pose} can generate pose-controllable talking face videos based on modularized audio-visual representation.  
\textbf{Wav2Lip}~\cite{prajwal2020lip} uses an encoder-decoder model learned via adversarial training to produce talking face videos. 
\textbf{MakeItTalk}~\cite{zhou2020makelttalk} leverages 3D landmarks to animate portrait images according to the input audio. 
\textbf{ATVGnet}~\cite{chen2019hierarchical} takes advantage of 2D landmarks to generate talking face videos from the input audio and an identity frame.
For more comparison settings, please refer to our supplementary document.

\subsection{Evaluation Metrics}
\textbf{Visual Quality Metrics.}
We use Peak Signal-to-Noise Ratio (\textbf{PSNR}) and Structured similarity (\textbf{SSIM}) \cite{wang2004image} to measure the similarity between generated and ground-truth images. Learned Perceptual Image Patch Similarity (\textbf{LPIPS}) \cite{zhang2018unreasonable} and Fréchet Inception Distance (\textbf{FID}) \cite{heusel2017gans} are employed to measure the feature-level similarity between generated and ground-truth images. We use the cosine similarity (\textbf{CSIM}) of identity vectors extracted by the face recognition network ArcFace~\cite{deng2019arcface} to evaluate the identity preservation ability.
\vspace{3mm}

\textbf{Lip Synchronization Metrics.}
Following \cite{xie2021towards}, we compute the normalized lip landmarks distance (\textbf{LipLMD}) between generated images and ground-truth images to evaluate the lip synchronization ability. 
When no ground-truth images exist, like in the video dubbing task, we employ SyncNet~\cite{chung2016out} to extract features from generated images and corresponding audio data and then calculate the \textbf{SyncScore}~\cite{chung2016out} between the two kinds of features. To a certain degree, this metric can convey how the lip shapes in generated images are coherent with the input audio.

\subsection{Quantitative Comparison}
We conduct a quantitative comparison of reconstruction and dubbing settings. In the reconstruction setting, we input the original audio to reconstruct the talking face videos. In the dubbing setting, the input audio comes from another video. Note that there is no ground truth for dubbed videos. Thus, quantitatively evaluating the generated results under the dubbing setting is challenging.

\textbf{Talking Face Reconstruction.}
~The quantitative evaluation results are reported in Table~\ref{tab:table1}. 
It can be seen that our method achieves the best performance on all visual quality metrics. 
Especially in identity preservation, our method is significantly better than other methods by large margins on the CSIM metric. 
The CSIM of our method is 10.09\% and 8.05\% larger than that of the second-best method, Wav2lip, on LRS2 and LRS3, respectively.
The reason is that our method can effectively leverage multiple reference images as well as landmark priors to provide more personalized facial attributes of a specific speaker, while other methods tend to generate the average facial attributes of the training dataset. 
Moreover, the LipLMD of our method is much smaller than those of other methods. This means that our method can generate more accurate facial landmarks as intermediate representation and the lip movement in our results is better synchronized with the input audio.

\textbf{Video Dubbing.} 
As shown in Table~\ref{tab:table1}, the SyncScore of our method is comparable to that of PC-AVS~\cite{zhou2021pose}, which adopts specific contrastive learning strategies to pull close the visual and auditory features. Meanwhile, our method achieves a much higher SyncScore than other methods.
This also validates that the lip shape generated by our method is well synchronized with the audio data.

\subsection{Qualitative Comparison}


\begin{table}[h]
\vspace{-10pt}
\fontsize{7.5}{9} \selectfont
\centering
\resizebox{0.9\columnwidth}{!}{%
\begin{tabular}{@{}c|c|c|c@{}}
\toprule
Method     & Image Quality & Lip Synchronization & Identity Preservation \\ \midrule
ATVGnet \cite{chen2019hierarchical}    & 1.78          & 2.42                & 2.22               \\
Wav2lip \cite{prajwal2020lip}    & 2.74          & 4.14                & 3.68               \\
MakeItTalk \cite{zhou2020makelttalk} & 2.94          & 1.97                & 3.10               \\
PC-AVS \cite{zhou2021pose}     & 2.98          & 3.47                & 2.94               \\
EAMM \cite{10.1145/3528233.3530745}       & 2.33          & 1.91                & 2.46               \\
Ours       & \textbf{4.40} & \textbf{4.56}       & \textbf{4.54}      \\ \bottomrule
\end{tabular}

}
\vspace{-5pt}
\caption{User study about video generation quality.}
\label{tab:user_study}
\vspace{-12pt}
\end{table}

\textbf{User Study.}~To validate the effect of our method qualitatively, we conduct a user study where 25 participants are invited to evaluate the generated videos.
Five videos are selected from each dataset for evaluation.
The input audio is chosen following the reconstruction setting.
Each participant is asked to rate the generated video  from 1 to 5 on three terms, including image quality, lip synchronization, and identity preservation. A higher score indicates a better result.
The mean opinion scores (MOS) are presented in Table \ref{tab:user_study}.
As can be observed, our method receives a better evaluation from participants than other methods. 
Specifically, the MOS of our method is 47.6\%, 10.1\%, and 23.4\% higher than that of the second best method on image quality, lip synchronization, and identity preservation, respectively.

\textbf{Visualization of Generated Images.} Three examples generated under the talking face reconstruction setting are presented in Fig.~\ref{fig:Qualitative}. Compared with other methods, our method can produce images that are visually closer to the ground truth.  
There also exist relatively less artifacts in our results.
PC-AVS~\cite{zhou2021pose}, Wav2Lip~\cite{prajwal2020lip}, MakeItTalk~\cite{zhou2020makelttalk}, and ATVGnet~\cite{chen2019hierarchical} fail to generate images that have consistent mouth shapes with the ground-truth images.
EAMM~\cite{10.1145/3528233.3530745} and PC-AVS tend to produce face images with moderate differences from ground truth images. For example, the eyes and wrinkles are apparently different from those in the ground truth images.
Besides, the results of Wav2Lip and ATVGnet are a little blurry.





\subsection{Ablation Study}
In this section, we conduct ablation studies on the LRS2 \cite{Afouras18c} dataset to validate the effect of core components in our method and the performance gain derived by multiple reference images.

\begin{table}[htb]
\fontsize{7.5}{9} \selectfont
\centering
\vspace{-10pt}
\resizebox{0.4\columnwidth}{!}{%
\begin{tabular}{@{}cc@{}}
\toprule
Method                      & Landmark Error \\ \midrule
w/ LSTM                 & 3.99           \\
\textbf{w/ Transformer} & \textbf{3.04}           \\ \bottomrule
\end{tabular}

}
\vspace{-8pt}
\caption{Ablation study on Transformer modules. }
\label{tab:ablation3}
\vspace{-10pt}
\end{table}

\textbf{Effectiveness of Transformer Encoder.}
To demonstrate the advantage of the transformer encoder used in landmark generator, we implement a variant of our method by replacing the transformer encoder with a commonly used bidirectional LSTM \cite{hochreiter1997long}. In this variant, reference embeddings are first averaged to form a global reference embedding as in \cite{xie2021towards}. 
Then, the LSTM module takes the concatenation of pose embedding, audio embedding, and global reference embedding as input to predict the landmarks. 
Landmark error is calculated to evaluate the accuracy of predicted landmarks (see supplementary material for more details). 
As shown in Table \ref{tab:ablation3}, our transformer-based landmark generator (`w/ Transformer') performs  better than the LSTM-based landmark generator (`w/ LSTM'). This is because the transformer module is more advantageous at modeling the temporal dependencies and the relationships between landmark and audio features.

\textbf{Effectiveness of Warping and Audio feature.}
We attempt to remove the warping mechanism in the alignment module and inject the misaligned reference images and features into the translation module. This forms a variant of our method indicated by `w/o warping'. As can be seen in Table \ref{tab:ablation_1}, it derives deteriorated visual quality metrics. 
Besides, the LipLMD metric also gets worse because misaligned reference images have negative impacts on inferring the lip shape. 
Moreover, without using audio feature to enhance the lip synchronization and mouth details in the translation module (i.e. `w/o audio'), both lip synchronization and visual quality metrics get worse. 
\begin{table}[t]
\fontsize{7.5}{9} \selectfont
\centering
 \setlength{\tabcolsep}{1mm}{
\begin{tabular}{@{}ccccccc@{}}
\toprule
Method        & PSNR↑          & SSIM↑           & LPIPS↓                  & FID↓           & LipLMD↓             & CSIM↑           \\ \midrule
w/o warping   & 32.46          & 0.9377          & 0.0323                  & 28.25          & 0.01357          & 0.6489          \\
w/o audio     & 32.71          & 0.9383          & 0.0317                  & 28.47          & 0.01334          & 0.6501          \\
\textbf{Ours} & \textbf{32.91} & \textbf{0.9399} & \textbf{0.0303} & \textbf{27.87} & \textbf{0.01293} & \textbf{0.6523} \\ \bottomrule
\end{tabular}
\vspace{-7pt}
}
\caption{Ablation study on audio features and reference image alignment in the landmark-to-video rendering model. 
}
\label{tab:ablation_1}
\vspace{-5pt}
\end{table}

\begin{table}[t]
\fontsize{7.5}{9} \selectfont
\centering
\vspace{-5pt}
 \setlength{\tabcolsep}{1mm}{
\begin{tabular}{@{}cccccc@{}}
\toprule
Reference Num & PSNR↑          & SSIM↑           & LPIPS↓          & FID↓           & CSIM↑           \\ \midrule
\textit{N=1}  & 32.91          & 0.9390          & 0.0313          & 28.55          & 0.6485          \\
\textit{N=5} & 32.96          & 0.9393          & 0.0312          & 28.35          & 0.6536          \\
\textit{N=10}  & 32.97          & 0.9394          & 0.0312          & 27.97          & 0.6529          \\
\textit{N=25} & \textbf{32.97} & \textbf{0.9394} & \textbf{0.0311} & \textbf{27.93} & \textbf{0.6546} \\ \bottomrule
\end{tabular}%
\vspace{-7pt}
}
\caption{Ablation study on the number of reference images.}
\label{tab:ablation_2}
\vspace{-18pt}
\end{table}

\textbf{Number of Reference Images.} We try to vary the number of reference images $N$ during inference to assess the performance gain derived by multiple reference images. 
The experimental results are provided in Table \ref{tab:ablation_2}.
We observe that using multiple reference images (5 to 25) gives rise to better metric values than using one reference image.


\section{Conclusion}
We propose a two-stage person-generic method for audio-driven talking face generation. 
First, we devise a novel transformer-based landmark generator to obtain accurate lip and jaw landmarks from audio. Then we align multiple reference images with the target expression and pose to provide more appearance prior for rendering face videos. Besides, acoustic features are utilized to enhance lip synchronization in the rendering stage. 
Extensive experiments show that our method can generate more realistic, lip-synced, and identity-preserving talking face videos than other person-generic methods.

\section{Acknowledgments}
This work was supported in part by the Guangdong Basic and Applied Basic Research Foundation (NO. 2020B1515020048), in part by the National Natural Science Foundation of China (NO. 61976250), in part by the Shenzhen Science and Technology Program (NO. JCYJ20220530141211024) and in
part by the Fundamental Research Funds for the Central Universities under Grant 22lgqb25.  This work was also sponsored by Tencent AI Lab Open Research Fund (NO.Tencent AI Lab RBFR2022009).

{\small
\bibliographystyle{ieee_fullname}
\bibliography{egbib}
}

\end{document}